\documentclass[letterpaper, 10 pt, conference]{ieeeconf}  

\IEEEoverridecommandlockouts                              
                                                          
\usepackage{amsmath}    			
\usepackage{amssymb}
\usepackage{amsfonts}

\usepackage{graphicx}  				

\usepackage[english]{babel}  
\usepackage[utf8]{inputenc}  	
\usepackage[T1]{fontenc}
\usepackage{ae,aecompl}     	

\usepackage{svg}
\usepackage{microtype}				
\usepackage[]{flushend}
\usepackage{caption}
\usepackage{cite}
\usepackage{subcaption}

\usepackage{tabularx}
\usepackage{booktabs}
\usepackage{enumerate}				
\usepackage[ruled,vlined]{algorithm2e}			
\usepackage{dirtree}					
\usepackage{hyperref}					
\usepackage{authblk}
\usepackage[left=1.69cm, right=1.69cm, top=2.01cm, bottom=1.52cm]{geometry} 
\graphicspath{{./figures/}}

\usepackage{url}
\usepackage{hyperref}
\usepackage{float}
\usepackage{todonotes}

\makeatletter
\newcommand{\removelatexerror}{\let\@latex@error\@gobble}

\makeatother

\author{Bruno Maric$^{*}$, Filip Zoric$^{*}$, Frano Petric and Matko Orsag%

\thanks{All authors are with the LARICS  Laboratory  for Robotics  and  Intelligent  Control  Systems, University of Zagreb, Faculty of Electrical Engineering and Computing, Zagreb 10000, Croatia
        {\tt\small (bruno.maric, filip.zoric, frano.petric, matko.orsag) at fer.hr}}

\thanks{* authors contributed equally.}%

\thanks{Ethics of the research conducted for this manuscript involving human subjects was approved by the Ethics Committee of the Faculty of Electrical Engineering and Computing, University of Zagreb, from 21. November 2023. }
\thanks{This work has been submitted to the IEEE for possible publication. Copyright may be transferred without notice, after which this version may no longer be accessible. }
}

\title{Comparative Analysis of Programming by Demonstration Methods: Kinesthetic Teaching vs Human Demonstration} 

\DeclareMathOperator*{\argmin}{arg\,min}

\begin{document}

\maketitle

\begin{abstract}
Programming by demonstration (PbD) is a simple and efficient way to program robots without explicit robot programming. PbD enables unskilled operators to easily demonstrate and guide different robots to execute task. In this paper we present comparison of demonstration methods with comprehensive user study. Each participant had to demonstrate drawing simple pattern with human demonstration using virtual marker and kinesthetic teaching with robot manipulator. To evaluate differences between demonstration methods, we conducted user study with 24 participants which filled out NASA raw task load index (rTLX) and system usability scale (SUS). We also evaluated similarity of the executed trajectories to measure difference between demonstrated and ideal trajectory. We concluded study with finding that human demonstration using a virtual marker is on average 8 times faster, superior in terms of quality and imposes 2 times less overall workload than kinesthetic teaching.  


\end{abstract}

\section{Introduction}

Robot manipulators can execute a wide variety of different tasks, and are a major factor in accelerating industrial automation. 
They excel in situations where their meticulously designed code does not require constant modifications. The lack of intuitive and fast programming methods has been the main deterrent to their application in agile production lines, where products and services change daily. Programming robots requires the use of trained robotics engineers, who often lack the practical experience for the task at hand.   

These challenges created a growing demand for collaborative robot manipulators \cite{grau2020_robots_in_industry} that facilitate safe human-robot interaction and easy robot programming. Since the advent of cobots, Programming by Demonstration (PbD) \cite{handbook} has gained widespread popularity in both academia and industry. PbD, also known as imitation learning or programming without coding \cite{lentini2020_programming_wo_coding}, aims to simplify robot deployment and eliminate explicit task programming. Demonstration can be divided into kinesthetic teaching, teleoperation, and passive observation \cite{Ravichandar2020}. Kinesthetic teaching refers to the force-guiding robot manipulator, teleoperation refers to the remote operation of the robot by operator, while in the passive observation scenarios the task is demonstrated by a human operator in manner which is the most natural for a given task. 

\begin{figure}
	\centering
		\includegraphics[width=0.95\linewidth]{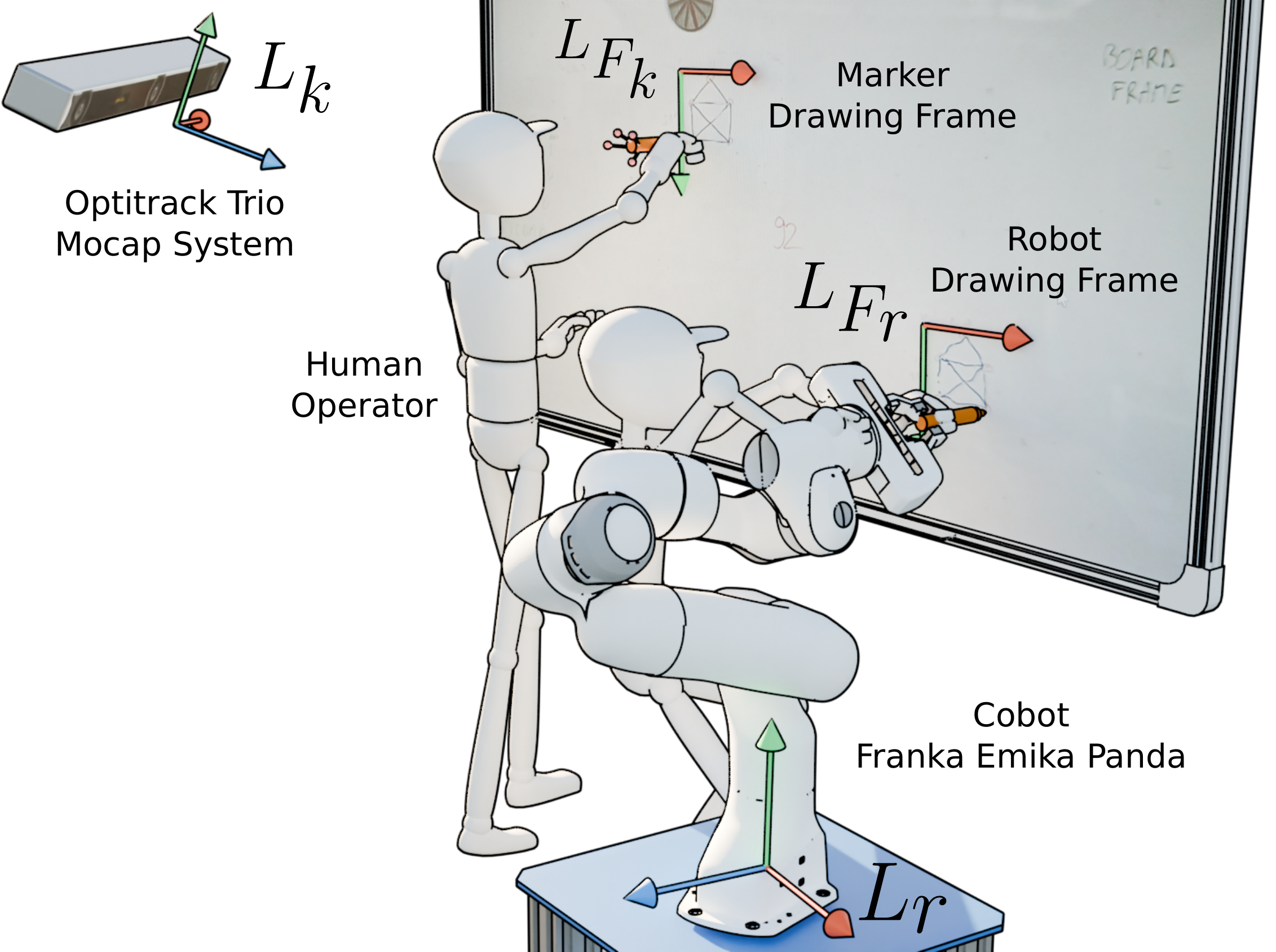}
	\caption{The experimental setup for the user-experience study comprised a motion capture system, specially developed virtual markers, and the collaborative robot \textit{Franka Emika Panda}. Participants were tasked to draw simple pattern using human demonstration with virtual marker (\textit{human operator on the left}) and kinesthetic teaching with cobot (\textit{human operator on the right}).}
	\label{fig:exp_sketch}
 \vspace{-0.7cm}
\end{figure}

With experience in various approaches to recording demonstrations for specific and delicate industrial applications, as well as for robot teleoperation \cite{ZoricH2AMI}, \cite{ZoricIROS}, this paper focuses on PbD using a specially developed virtual marker for passive demonstration. We argue that the adoption of newer and simpler interfaces for demonstrations can facilitate further utilization of the robot manipulators, be they collaborative or standard industrial robots capable of diverse tasks when programmed effectively.

To test our hypothesis, we compare passive demonstration using virtual marker with kinesthetic teaching using robot manipulator via a comprehensive user study. In the user study we estimated imposed workload and system usability, whilst measuring demonstration duration. We have designed a simple drawing task that involves reaching of several waypoints (connect the dots), path following between the waypoints and requires maintaining the contact with the surface at all times. The aforementioned aspects of the task encompass a wide range of currently manual operations in the industry, in tasks such as welding, sanding, cutting, engraving etc. The task is designed after a popular puzzle given to children, ensuring that all participants are familiar with the task, as is expected in the envisioned industrial setting, and that the only novelty for a participant in the study is the method for demonstration of the task.
\textit{As the main contribution of the paper we present a comprehensive analysis of the conducted comparative study that explores workload and usability differences between kinesthetic teaching and human demonstration. }


\section{Related work}
\label{section:related_work}

Programming by Demonstration (PbD) has gained significant popularity in both academia and industry over the last 30 years. As outlined in \cite{Zhang2016}, PbD involves two main phases: learning and representation. During the learning phase, demonstrations are collected and segmented into action representations, forming an assembly tree that indicates the order of actions. Different methods, such as kinesthetic teaching \cite{Tykal2016_Kinesthetic} or teleoperation \cite{Zhang2018_VR}, can be used for this phase. The representation phase is then followed with robot movement mapping and robot execution. Various representation approaches are available, spanning probabilistic models \cite{Calinon2009}, data-driven AI-based models \cite{Scherzinger2019Contact}, and the increasingly popular Dynamic Movement Primitives (DMP) \cite{pastor2011_ICRA_DMP} in recent years. Our paper contributes to this field by presenting and systematically comparing different demonstration techniques to determine the most effective way for demonstrations.

When it comes to interesting PbD applications, Wang et all. \cite{Wang2023} proposed novel approach for insertion task, where the robot is able to recreate  precise insertion task just by passively observing human doing it once. Visual servoing is utilized to enable human hand tracking. In the approach evaluated in this work, instead of visual servoing, the motion capture system is deployed to track the virtual marker, which enables recording and further processing of recorded motion, introducing flexibility and allowing for demonstrations from different operators. Authors in \cite{ajaykumar2023} propose PbD system that besides kinesthetic teaching incorporates different modalities that humans use when naturally communicating some physical task or a mission, such as gaze and speech. As authors report, using multimodal PbD can lead to overtrust and automation bias in the long term, which is why it makes sense to explore HRI through PbD with different modalities of kinesthetic teaching and human demonstration, but also to include other modalities that are synchronized with virtual marker motion, such as force or human pose measurements. In \cite{steinmetz2019} authors propose task level PbD which can be quite useful for the introduction of the collaborative robots in the small and medium sized SMEs. Main difference compared to our approach is that they use kinesthetic teaching coupled with on-line semantic skill recognition algorithm which enables untrained operators to demonstrate certain task, while we aim to capture the skill of a task expert.

\section{Virtual Marker}
\label{section:virtual_pen}


Building upon our previous work \cite{Maric_KDNO}, which introduced the concept of utilizing a motion capture system and a specially designed virtual marker for measuring geometrical relationships in 3D space, this study shifts focus towards the application of using virtual marker in Programming by Demonstration (PbD) scenarios. Contrary to standard collaborative robotics approach, where operator uses force to guide robot's flange, we strongly believe there are numerous different applications where such approach fails to capture true essence of the demonstrated skill. 

\subsection{Calibration methods}
\label{subsection:kalipen_calibration}

Given that the Optitrack motion capture system tracks the global pose $\mathbf{T}_W^I$ of the fiducials placed on top of the virtual marker in world frame, precise calibration of the pen tip becomes crucial. The calibration process involves deriving the relative transformation between the virtual marker tip and the Optitrack marker, represented as $\mathbf{T}_I^P$. Building upon that, the calculation of the global pose of the virtual marker tip $\mathbf{T}_W^P$ is obtained as:
\vspace{-0.1cm}
\begin{equation}
    \mathbf{T}_W^P = \mathbf{T}_W^I \cdot \mathbf{T}_I^P .
    \label{eq:tip_position}
\end{equation}

The virtual marker calibration process consists of two steps. The first step establishes the translational relation of the marker tip frame ($L_P$) with respect to the Optitrack fiducials frame ($L_I$), while in the second step the focus is on adjusting the orientation of the frame $L_P$.


\subsubsection{Position calibration} 
To calibrate the position of the marker tip, the virtual marker tip $\mathbf{T}_\mathrm{W}^{\mathrm{P}}$ is fixed at a specific point in the world reference frame $\mathbf{L}_W$, while the reflective fiducial part is moved and rotated throughout the space. The objective of optimization problem is defined as:
\vspace{-0.1cm}
\begin{equation}
  \mathbf{P}^\ast = \argmin\limits_{\mathbf{P}} \sum_{\mathit{i}=1}^{\mathrm{N}} \sum_{\mathit{j}=1}^{\mathrm{N}} ||{\mathbf{T}_{\mathrm{O}, \mathit{i}}^\mathrm{I} \cdot \mathbf{T}_\mathrm{I}^{\mathrm{P}} - T_{\mathrm{O}, \mathit{j}}^\mathrm{I} \cdot \mathbf{T}_\mathrm{I}^{\mathrm{P}}}|| ,
  \label{eq:marker_opti_pos}
\end{equation}
\noindent where the goal is to minimize the global distance discrepancies between the transformed points obtained using Eq. \ref{eq:tip_position} across the entire calibration dataset. Here $\mathbf{P}^\ast \in \mathbb{R}^{3 \times 1}$ presents optimal translation of the $\mathbf{T}_\mathrm{I}^{\mathrm{P}}$.

\subsubsection{Orientation Calibration} 
Orientation of the virtual marker can be performed using a specially designed tool, which has several holes, each with a predetermined orientation relative to the motion capture system's reference frame. The calibration for the orientation is also formulated as an optimization problem, with the goal to modify the rotational part of $\mathbf{T}_\mathrm{I}^{\mathrm{P}}$ to align the measured vector $z_p$ with the hole orientation. More details on the calibration procedure, different virtual and calibration tools are available in \cite{Maric_KDNO}.

\subsection{Deployment} 

Following the proper calibration procedures, the virtual marker is capable of measuring and marking points in the global reference frame with submillimeter accuracy and orientation precision of up to 1 degree in approach axis as shown in \cite{Maric_KDNO}. This level of precision enables the marker to be used in various applications, especially those that require precise end-effector positioning over extended period of times, such as welding, sanding and drilling. As such, it is valuable for the efficient configuration of robotic systems, enabling the capture of robot tasks and surroundings without the need for extensive measurements or modeling. The recordings obtained with marker can be synchronized and augmented with other signal modalities of importance for a given task, such as force/torque measurements \cite{maric_case_drilling}. 


\section{Experiment Methodology}
\label{section:methodology}


To assess the ergonomics and user-friendliness of the virtual marker in PbD tasks, we conducted a user study. Participants were tasked with demonstrating drawings using both the virtual marker and by guiding the flange of the collaborative manipulator \textit{Franka Emika Panda}. Our primary objective was to validate our claims regarding the potential of this new Human-Robot-Interface (HRI) to enhance the adoption of robot manipulators, even among unskilled robot operators. 

\subsection{Hardware setup}
In our experimental setup, we used the virtual marker tool with the \textit{Optitrack} mocap system, a collaborative manipulator \textit{Franka Emika Panda}, force/torque sensor and a whiteboard. Recognizing the significance of the contact force in the demonstration, we strategically positioned the force-torque sensor beneath the whiteboard. This setup allows us to concurrently track the contact force along with tracing the virtual marker, resulting in the following measurement vector:
\vspace{-0.2cm}

\begin{equation}
     \mathbf{p} =  \left[t \quad x \quad y \quad z \quad F_z \right] \in \mathbb{R}^{1 \times 5},  
     \label{eq:point_P}
\end{equation}
\noindent where the virtual marker tip position at time $t$ is denoted as $\mathbf{p}_p = \left[ x \quad y \quad z \right] \in \mathbb{R}^{1 \times 3} $, while the $F_z \in \mathbb{R}$ represents the contact force between the marker and the whiteboard in one axis at time $t$. 

To simulate demonstrations of tasks where the effect of the tool can be observed during demonstration and those where such demonstration is not feasible (safety, delicacy of the part etc), we used two marker versions: one that leaves a trace on the whiteboard, and the other one that does not. Both marker versions had the same physical dimensions, differing only in their tracing capability.

The experimental setup is depicted in Fig. \ref{fig:exp_sketch}. Prior to the experiment, we conducted calibration for the virtual markers, as outlined in Sec. \ref{subsection:kalipen_calibration}, resulting in two homogeneous transformation matrices: $\mathbf{T}_{O_k}^{P_t}$ and $\mathbf{T}_{O_k}^{P_n}$. A similar calibration procedure was followed for the robot flange to tool transformation $\mathbf{T}_{F}^{P}$, with the difference of using the flange pose derived from the robot's Forward Kinematics instead of the \textit{Optitrack} measurement ${\mathbf{T}_W^I}$.

\subsection{Drawing frames}

To standardize user drawings created with both the virtual marker and the robot, and facilitate uniform post-analysis, we introduced drawing frames. These frames were selected and marked on the whiteboard, and a 3D printed template was aligned within the frame to mark five waypoints for the drawing. This ensured consistency and comparability across all participant drawings. The virtual marker drawing frame is labeled as $L_{F_k}$, while the robot drawing frame is denoted as $L_{F_r}$.

The drawing frames were marked using both the virtual marker tip $\mathbf{T}_W^P$ and the robot tip $\mathbf{T}_B^P$. To explain the methodology, let us denote the tool tip pose as $\mathbf{T}_{O_x}^P$, its translational component as $\mathbf{p}_{O_x}^P$, and the drawing frame as $L_{F_x}$. The drawing frame $L_{F_x}$ was localized using three points to form a freely chosen right triangle, marked in the following sequence: 1) $\mathbf{T}_{F_x}^{P_1}$ on the $+x$ axis; 2) $\mathbf{T}_{F_x}^{P_2}$ at the origin; and 3) $\mathbf{T}_{F_x}^{P_3}$ on the $+y$ axis. Following this, coordinate frame unit vectors are derived using:

\begin{equation}
    \begin{aligned}
        \mathbf{x}_{O_x}^{F_x} = \frac{\mathbf{p}_{O_x}^{P_1} - \mathbf{p}_{O_x}^{P_2}}{|| \mathbf{p}_{O_x}^{P_1} - \mathbf{p}_{O_x}^{P_2} || },  
        \mathbf{y}_{O_x}^{F_x} = \frac{\mathbf{p}_{O_x}^{P_3} - \mathbf{p}_{O_x}^{P_2}}{|| \mathbf{p}_{O_x}^{P_3} - \mathbf{p}_{O_x}^{P_2} || }. 
    \end{aligned}
    \label{eq:kalipen_frame_ident}
\end{equation}
Having calculated unit vectors $\mathbf{x}_{O_x}^{F_x} \in \mathbb{R}^{3 \times 1}$ and $\mathbf{y}_{O_x}^{F_x} \in \mathbb{R}^{3 \times 1}$, remaining one is calculated as: $\mathbf{z}_{O_x}^{F_x} = \mathbf{x}_{O_x}^{F_x} \times \mathbf{y}_{O_x}^{F_x}$. Following that, the matrix of homogeneous transformation of the frame $T_{O_x}^{F_x}$ is derived as follows:
\begin{equation}
    T_{O_x}^{F_x} = 
    \begin{bmatrix}
    \begin{matrix}
        {\mathbf{x}_{O_x}^{F_x}} & {\mathbf{y}_{O_x}^{F_x}} & {\mathbf{z}_{O_x}^{F_x}}
    \end{matrix} & \mathbf{p}_{O_x}^{P_2} \\
    \mathbf{0} & 1
    \end{bmatrix}
\label{eq:kalipen_frame_ident_2}
\end{equation}
\noindent where $\mathbf{p}_{O_x}^{P_2} \in \mathbb{R}^{3\times 1}$ is translational component of the measured point $\mathbf{T}_{0_x}^{P_2}$.


While drawing, the tool tip pose is tracked in the global frame of the mocap system or the robot base link, generally expressed as ${\mathbf{T}_{O_x}^{P}}_{, i}$. To transform the obtained points $ {\mathbf{T}_{O_x}^P}_{, i}$ into the drawing frame, the following transformation is applied:

\begin{equation}
    {\mathbf{T}_{F_x}}_{, i} = {\mathbf{T}_{O_x}^{F_x}}^{-1} \cdot {\mathbf{T}_{O_x}^P}_{, i}. 
    \label{eq:kalipen_point_transformation}
\end{equation}

%
%

\begin{figure*}[]
    \centering
    \includegraphics[width=0.99\linewidth]{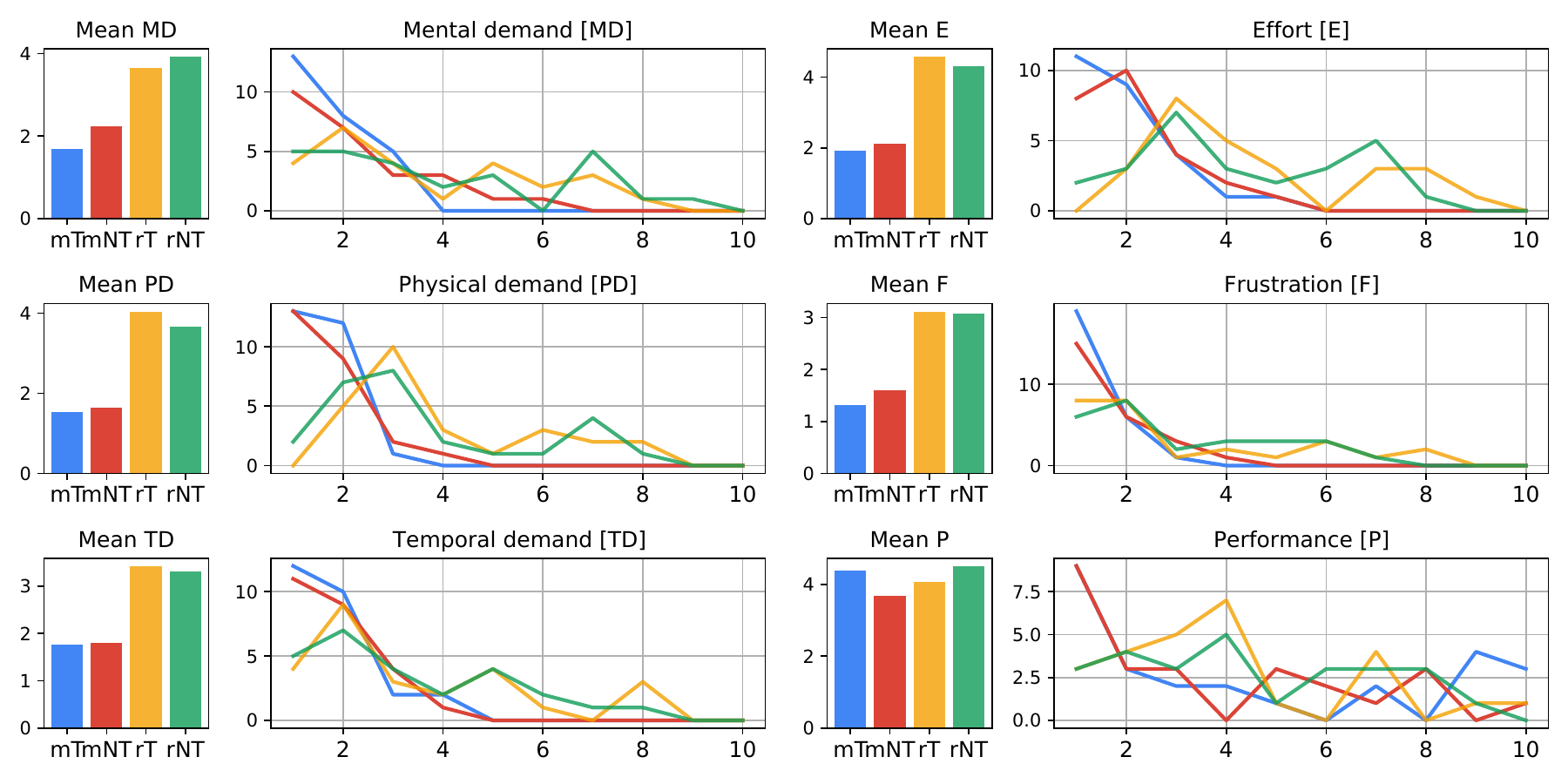}
    \caption{NASA raw task load index (rTLX) questionnaire results. Bar charts show mean ratings of all participants. Line plots show frequency of certain rating for each task in the data acquired. For the workload measurement, \textit{lower rating is better}. Graphs show that virtual marker introduces significantly less workload compared to the robot guidance.}
    \label{fig:nasa_rtlx}
\end{figure*}

\subsection{Task and Population samples}
\label{subsection:pop_sample}

To show difference between demonstration methods for the same task, we divided participants in 8 different groups to mitigate influence of knowing tasks beforehand as explained in the section \ref{section:experiments}. To each study participant we showed  a \href{https://docs.google.com/presentation/d/e/2PACX-1vSHSBL3lApqPiP_uIFSpdMoo2_hq4WFV6JP91JyYrnFCWOVkF-75GNw44UytkGG7qBDYPad65uefwsR/pub?start=false&loop=false&delayms=5000}{video} to demonstrate how to hold the marker, and how to guide the robot manipulator. Each user had to complete the drawing in four different ways: A) virtual marker that leaves trace, B) virtual marker that leaves no trace, C) trace-leaving marker on a cobot, D) traceless marker on a cobot. 

There was 24 study participants. Participants ranged in age from 22 - 28. Gender information is omitted because it doesn't affect study outcome. Participants were divided into 8 groups, and each group had three participants that executed tasks in different ordering (ABCD, ABDC, BACD, BADC, CDAB, DCAB, CDBA, DCBA). 

\section{Experimental results}
\label{section:experiments}


Every participant executed 4 different tasks in certain ordering (e.g. ABCD), and after each task, participant was prompted to fill out NASA raw Task Load Index (rTLX) and the system usability scale (SUS) test, as explained in-depth in subsections \ref{subsection:nasa_rtlx} and \ref{subsection:sus}. Following all participants, task evaluation was performed offline.  



\subsection{NASA Raw Task Load Index}
\label{subsection:nasa_rtlx}


NASA task load index \cite{NASA_TLX1}, \cite{NASA_TLX2} is measurement developed by NASA to assess operator's workload when instructing or operating a machine. It consists of two parts. First part is used to assess subjective importance of each workload category. Second part is used to asses each workload category. We omitted first part of the assesment, and employed only NASA raw task load index \cite{Bustamante2008}. After each task, participants had to estimate each of the following workload categories: physical demand (PD), mental demand (MD), frustration (F), performance (P), temporal demand (TD) and effort (E). Compared to the original NASA TLX we have reduced rating scale from 0-21 to 0-10 because it is simpler for users to populate such questionnaire, and as such, does not affect end results. Digital form used to collect data can be found \href{https://docs.google.com/forms/d/1vjNKyp12z_yzD3WH1ztPmWyXYo3LIB--DrsxUc5SNhQ}{here}. In the Fig. \ref{fig:nasa_rtlx} it can be seen that almost all of the workload categories (except performance), when averaged across all study participants were significantly lower for the virtual marker than for the robot manipulator. Such results indicate that virtual marker induces  less operator workload compared to the kinesthetic teaching. 


\subsection{System usability scale}
\label{subsection:sus}

\begin{figure}
\centering
\includegraphics[width=0.95\linewidth]{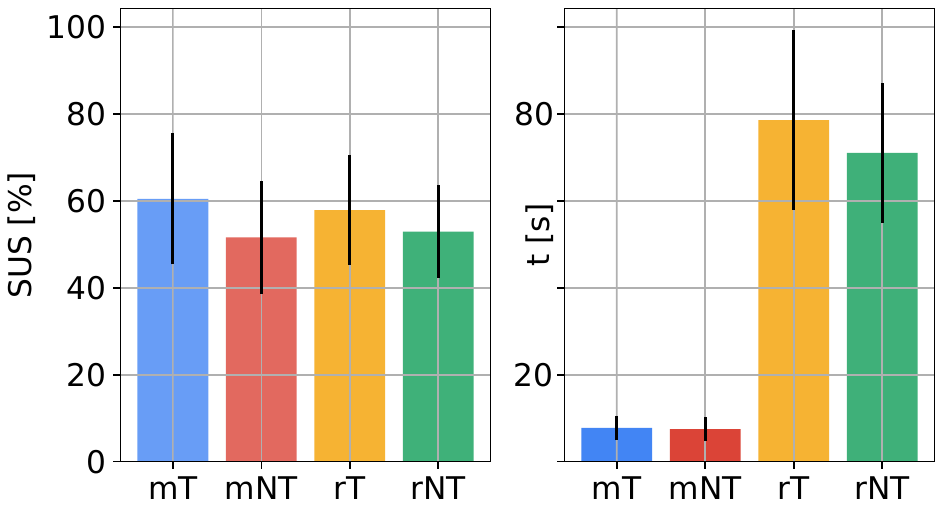}
\caption{Left plot shows mean System usability score (SUS) (\textit{higher is better}).  Right plot shows saverage time for different programming modality (\textit{lower is better}). Users determined that virtual marker that leaves trace is best in terms of usability. From the demonstration time perspective, using virtual marker speeds up demonstration process on average, eight times.}
\label{fig:sus_results}
\vspace{-0.5cm}
\end{figure}

In-depth explanation of the system usability scale can be found in \cite{Brooke1995}. Usability of a system, device or procedure can not be defined outside of context. Context is defined by: users of the system, tasks performed by it, and and the environment in which it will be used. As we already mention, virtual marker demonstration aims to alleviate programming burden from unskilled robot operators in task demonstration. Through our experimental procedure we recreated such environment by no interference with the participant during their task demonstration. Averaged results across participants of the SUS as well as average demonstration time for programming modality can be found in  Fig. \ref{fig:sus_results}.  

\subsection{Task evaluation}

Alongside the user experience evaluation, we performed a drawing task assessment. Since the task involved connecting marked points with straight lines, we aim to evaluate the deviation of demonstrated trajectories from the ideal lines connecting these marked waypoints. Initially, we transformed all recodings in it's drawing frame, using the Eq. \ref{eq:kalipen_point_transformation} and obtained frame transformations $L_{F{k}}$ and $L_{F{r}}$. Then we segmented all displayed trajectories into eight sections. Given the variations in demonstration speeds, leading to differing numbers of points in the demonstrated trajectories, a prerequisite for conducting any analysis on these points was the resampling of demonstrated trajectories. This approach ensures an equal number of trajectory points across participants and drawing cases.

\begin{figure}[]
    \centering
    \begin{subfigure}{0.99\linewidth}
          \centering
          \includegraphics[width=.99\linewidth]{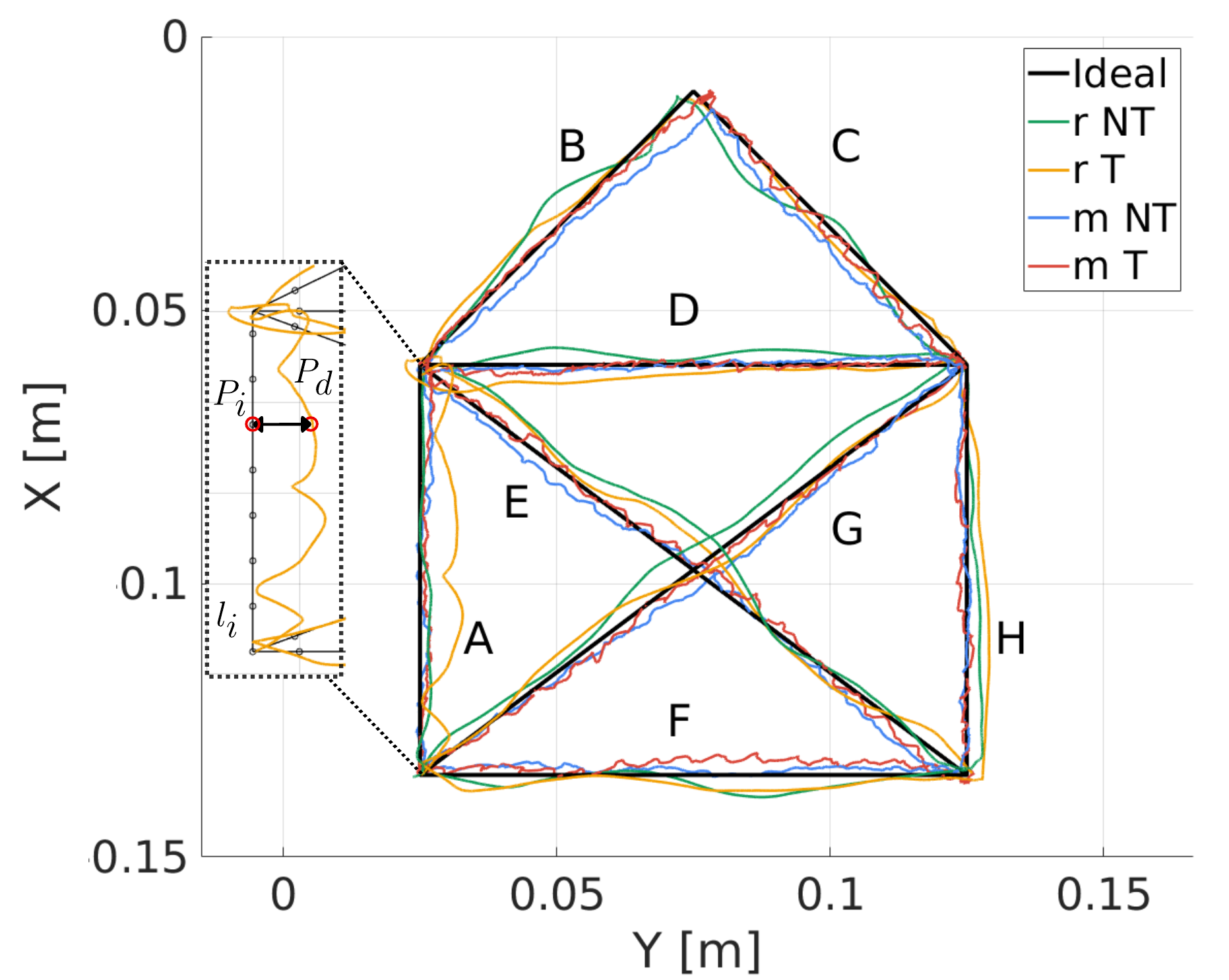}
          \caption{}
          \label{fig:TaskEvaluation_trajectory_force_2D}
    \end{subfigure} 
    
    \begin{subfigure}{0.99\linewidth}
          \centering
          \includegraphics[width=.99\linewidth]{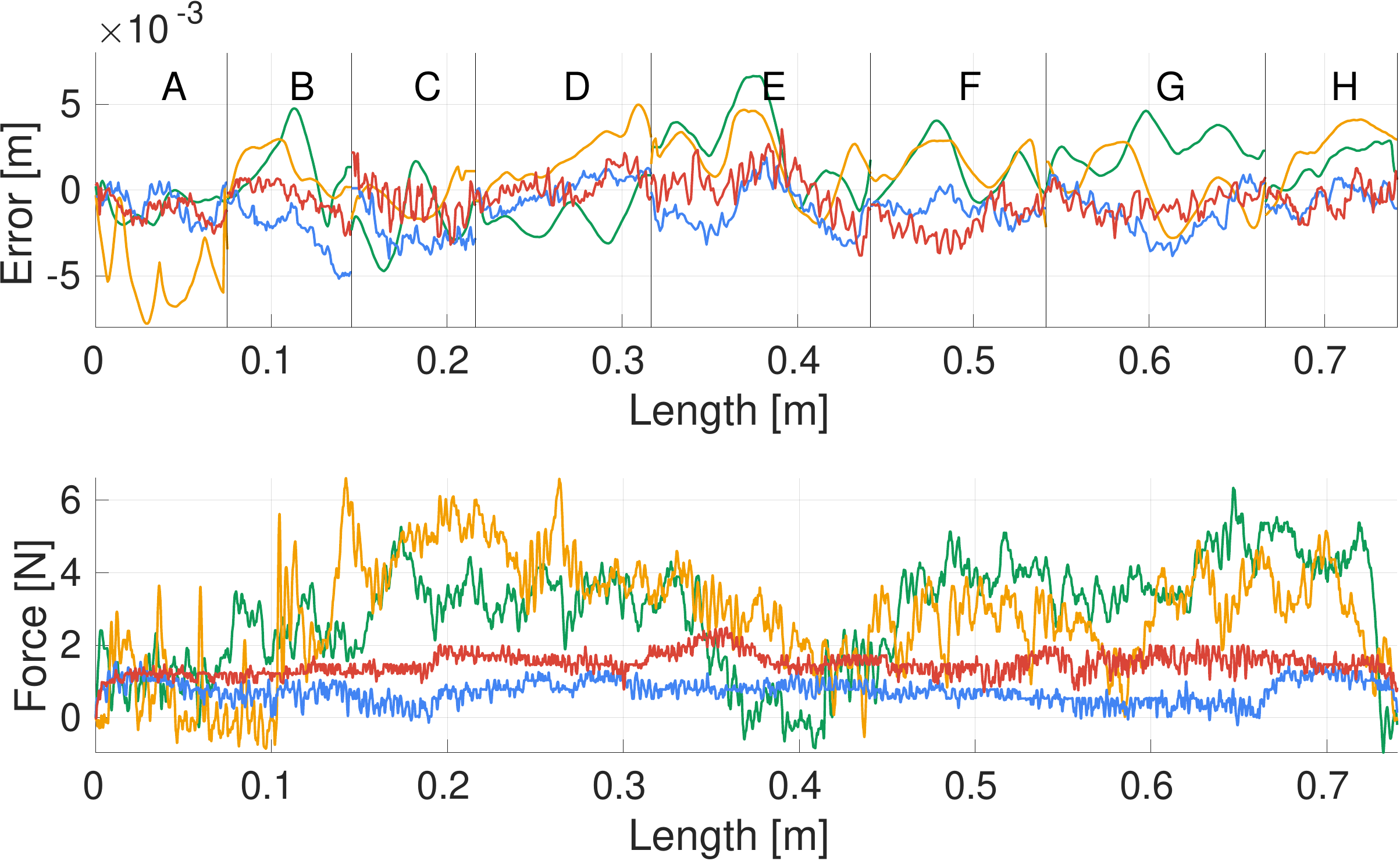}
          \caption{}
          \label{fig:TaskEvaluation_trajectory_force_1D}
    \end{subfigure}

    \caption{Top: the recordings of a single participant with the cobot (no-trace in green and trace in yellow) and the virtual marker (no trace in blue and trace in red). The detail demonstrates the sampling of demonstration, with marked point $P_i$ on ideal line $l_i$ and point $P_d$ on demonstrated trajectory. Middle: the distance error between the demonstrated trajectories and the ideal trajectory for each segment (A-H). Bottom: contact force for each demonstrated trajectory along the drawing path.}
    \label{fig:TaskEvaluation_trajectory_force}%
\vspace{-0.5cm}
\end{figure}

In the resampling process, the ideal segment line $l_{i}$ was uniformly divided into a consistent number of points per segment. Each point $P_i$ on $l_{i}$ was then paired with a corresponding point $P_d$ on the demonstrated trajectory, orthogonal to $l_{i}$ and $P_i$, as shown in Fig. \ref{fig:TaskEvaluation_trajectory_force} in the upper graph detail. The orthogonal projection of $P_d$ to $l_{i}$ represents the distance error from the demonstrated to the ideal trajectory. This process was conducted for each participant and drawing case, including pairing contact forces exerted between the marker and the whiteboard. Fig. \ref{fig:TaskEvaluation_trajectory_force} illustrates the prepared trajectories for a single participant, showing ideal and demonstrated trajectories (robot and marker), distance errors, and contact forces on the whiteboard.


\begin{figure}
\centering
\includegraphics[width=0.99\linewidth]{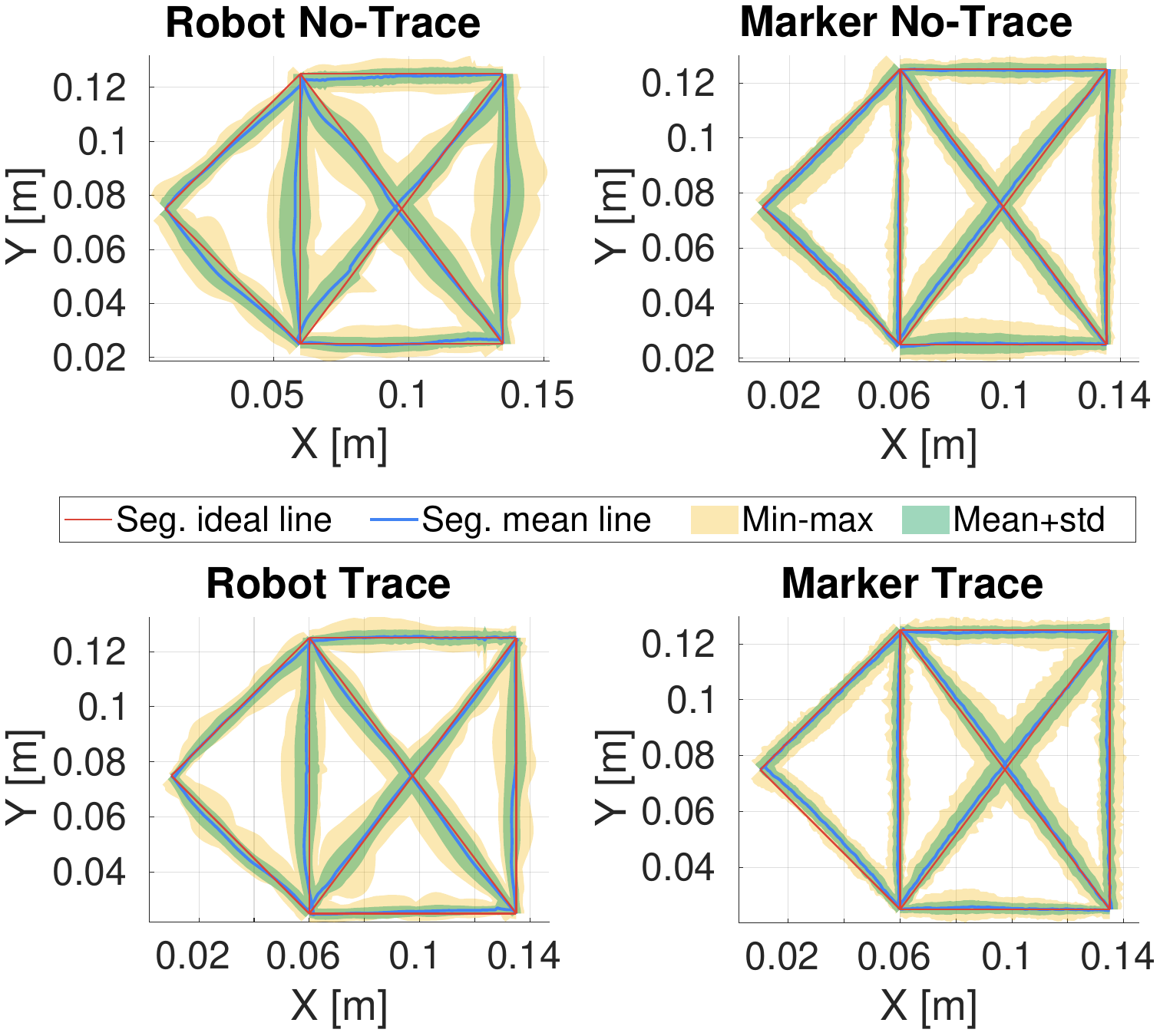}
\caption{The mean value of all demonstrations (blue), compared to the ideal trajectory highlighted in red. The mean value is derived from distances between the points of trajectories and their respective orthogonal projections onto the ideal segment line. Yellow indicates the envelope of trajectory values, while the green area portrays the standard deviation from the mean trajectory. }
\label{fig:TaskEvaluation_trajs}
\vspace{-0.2cm}
\end{figure}

In order to overlap the trajectories of all participants for each test case, we derived an envelope encompassing all trajectories per segments, as illustrated in Fig. \ref{fig:TaskEvaluation_trajs}. This depiction highlights that the kinesthetic demonstration spans a broader area compared to the ideal waypoints. Additionally, we computed the mean trajectory, obtained from the distance errors from the demonstrated to the ideal trajectory. Alongside the mean trajectory, we presented the standard deviation, representing the dispersion around the mean trajectory.

\begin{figure}
\centering
\includegraphics[width=0.99\linewidth]{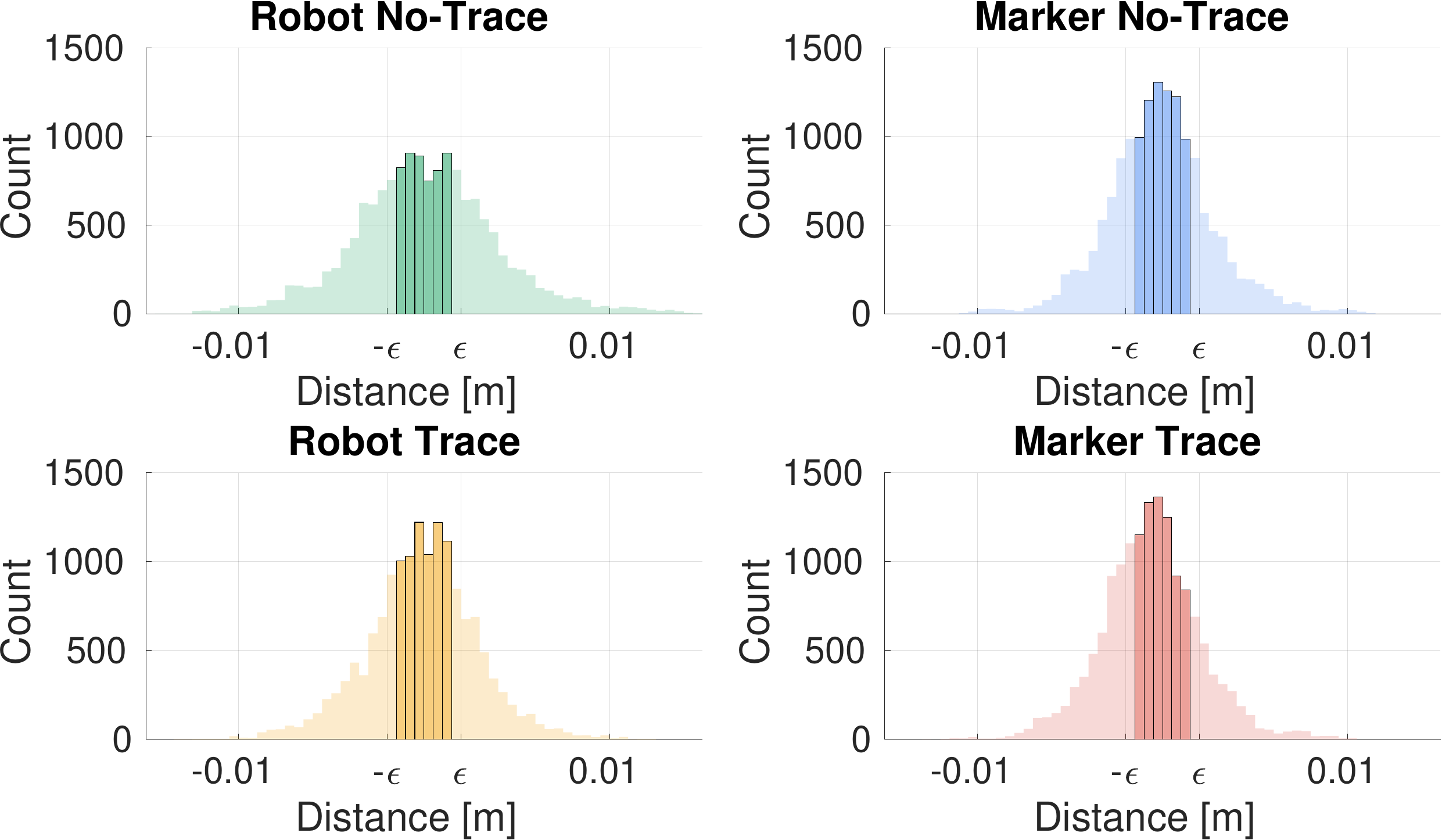}
\caption{Distribution of distances of trajectory points from the ideal segment's lines, accompanied by the corresponding count of such points. The narrow $\epsilon$ area, indicating where the majority of well-demonstrated trajectory points are expected, is highlighted.}
\label{fig:TaskEvaluation_histogram}
\vspace{-0.4cm}
\end{figure}

To substantiate that trajectories demonstrated with the virtual marker are more compact, closely aligned with the ideal lines, and exhibit a smaller dissipation rate, we employ additional metrics. The histogram depicted in Fig. \ref{fig:TaskEvaluation_histogram} illustrates the groups of distances of the trajectory points from the ideal line, with the corresponding count of such points. Additionally, we highlight the \textit{Epsilon} area in blue, representing a narrow $3$ mm wide zone around the ideal lines, where we anticipate the majority of trajectory points to fall. It is evident that for virtual marker demonstrations, a greater number of trajectory points lie within the \textit{Epsilon} area.

\begin{figure}
\centering
\includegraphics[width=0.99\linewidth]{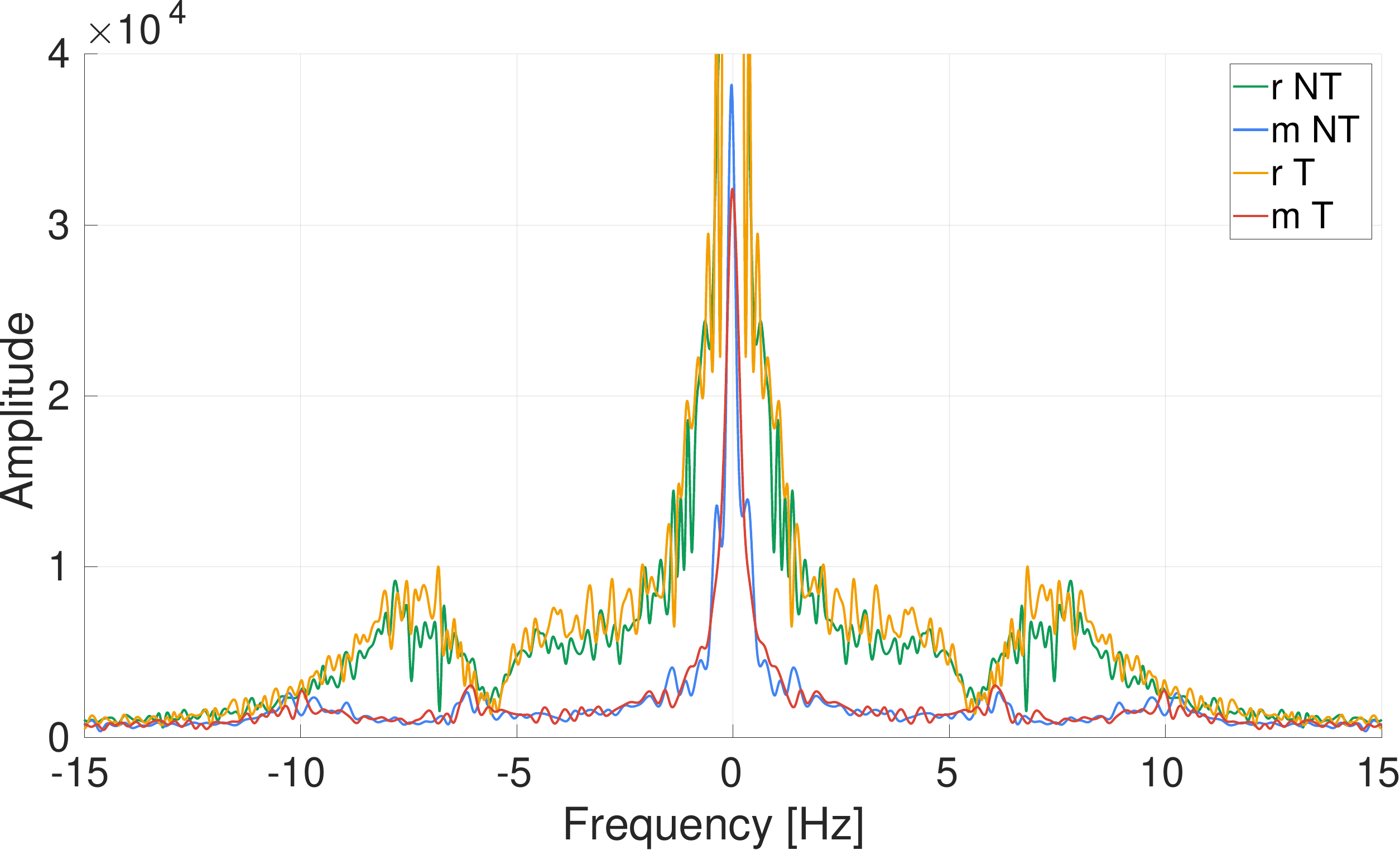}
\caption{Fast-Fourier-Transformation of force signals. Demonstrations where marker does not leave trace are shown in green (robot) and blue (marker), while yellow (robot) and red (marker) denote demonstrations with the marker that does leave trace. }
\label{fig:TaskEvaluation_fft_histogram}
\vspace{-0.6cm}
\end{figure}

Two observations stand out from the contact force $F_z$ during the demonstration, as depicted in Fig. \ref{fig:TaskEvaluation_trajectory_force}. Firstly, the force amplitude is higher in the robot demonstration, and secondly, the force signal is significantly more variable in such cases. In contrast, the contact force signal in virtual marker demonstrations has a lower amplitude but is more consistent. The variation in the force signal is particularly interesting in our experiment. To investigate further, we conducted Fast Fourier Transformation (FFT) on the force signal $F_z$ for each demonstration. In Fig. \ref{fig:TaskEvaluation_fft_histogram}, the results of the FFT are presented in the form of a histogram. The y-axis represents the count of frequencies found in the force signal, with amplitudes higher than a set threshold . It is evident that force signals from robot demonstrations encompass a broader range of frequencies, observed in both markers that leave a trace and those that do not. This can be attributed to the challenges associated with guiding the robot flange, often resulting in the loss and re-establishment of contact between the marker and the whiteboard.

\section{Discussion}
\label{section:discussion}

In the conducted user-experience survey, it is apparent that demonstrating with the virtual marker induces significantly less operator workload compared to directly guiding the cobot (see Fig. \ref{fig:nasa_rtlx}). Interestingly, the performance indicator averaged across participants is similar across all demonstration methods (see lower right Performance (P) plot in Fig. \ref{fig:nasa_rtlx}). Such results are unexpected, especially considering other workload categories (PD, TD, MD, E, F) show clear trends toward demonstration with marker as being preferred. We suspect that such discrepancy arises from the participant's natural bias incorporated by the rating system learned throughout education. In the local educational system highest grade (5 means best performance), and lowest grade (1 means worst performance or failure). Therefore we believe that some participants are habitually giving high ratings for good performance - disregarding instructions in the questionnaire. Nonetheless, participants did find the virtual marker system more useful in PbD applications compared to the cobot (see Fig \ref{fig:sus_results}). 

The task evaluation showed that trajectories demonstrated with the virtual marker were closer to the ideal task, exhibiting lower error rates and less variation than those demonstrated with the cobot (see Fig. \ref{fig:TaskEvaluation_trajs} and Fig. \ref{fig:TaskEvaluation_histogram}). Furthermore, demonstrations with the cobot exerted higher contact force with greater variations compared to virtual marker demonstrations (see Fig. \ref{fig:TaskEvaluation_fft_histogram}). All these findings suggest that directly guiding the cobot to demonstrate a specific task can be difficult, and the quality of the demonstration may suffer from the operator being outside their familiar environment and comfort zone. This situation may lead to demonstrations that fail to capture the true essence of the demonstrated motion.

Possible study limitation arises from the small diversity of the study population. It is not known to which extent age and gender affect subjective workload estimate. In our case, small age diversity in the study population contributes to the research as we can easily compare workload imposed by different PbD techniques, rather than exploring age effect on the felt workload. Additionally, although mitigated with the design of a very familiar task, the expertise of the demonstrator was not controlled variable and it's effect on the workload and demonstration method preference cannot be evaluated with this study.     

To contextualize these results in terms of industry, we compare the reported outcomes of two demonstration approaches in the case of robotic deep-micro-hole drilling of moulds used in the glass manufacturing industry. The first approach, as presented by Ochoa et al. \cite{ochoa2021impedance}, involves operators guiding the robot's end-effector during the demonstration. In contrast, our recent work \cite{maric_case_drilling} focuses on capturing the operator skill with the virtual marker, analyzing both the forces exerted and position profiles of the drilling tool with respect to the mould. Our study in \cite{maric_case_drilling} demonstrates significant performance differences compared to \cite{ochoa2021impedance}, underscoring the importance of keeping operators within their comfort zone for the precise capture and successful robotic reproduction of expert skills.

\vspace{-0.05cm}
\section{Conclusion}
\label{section:conclusion}
\vspace{-0.05cm}
In this work we presented a novel concept of using virtual marker in Programming by Demonstrations use cases. In the beginning we derived hardware setup and developed  calibration methods that facilitated the utilization of the virtual marker for virtual 3D mapping. Emphasizing the importance of keeping the operator within their comfort zone during demonstrations, we conducted a user-experience survey. Participants demonstrated the task using both the virtual marker and the robot, and their experiences were evaluated through workload and system usability questionnaires. 

Results indicated that participants found the virtual marker to be less physically and mentally demanding while maintaining high performance rates and demonstrating lower frustration and effort compared to using the collaborative robot. 
We also found that pre-existing cultural bias regarding rating procedures, can influence user study. We recommend everyone conducting user study to explicitly state possible biases to participants in order to make them fully aware of them and prevent them from affecting end result. 
Usability tests showcased the potential of the virtual marker in PbD tasks, surpassing direct guiding of the cobot. Besides that, virtual marker speeds up demonstration process on average, 8 times. Such speed up can bring various benefits to the agile production lines, limiting downtime, and increasing company profits. 
When it comes to the demonstration analysis, the results revealed that demonstrations using the virtual marker were more consistent, significantly reducing errors in drawing compared to demonstrations with the robot. This work demonstrates the promising applications and advantages of integrating a virtual marker in PbD scenarios, offering a more user-friendly and efficient alternative for collaborative robotics. In future, we aim to evaluate novel PbD modalities based on human pose estimation inspired by \cite{ZoricH2AMI}, \cite{ZoricIROS}, and exploring how to utilize such interface in real world environment for the HRI.

\addtolength{\textheight}{-5cm}   



\section*{ACKNOWLEDGMENT}
This research was a part of the scientific project Strengthening Research and Innovation Excellence in Autonomous Aerial Systems - AeroSTREAM \cite{AEROSTREAMweb} supported by European Commission HORIZON-WIDERA-2021-ACCESS-05 Programme through project under G. A. number 101071270. 
The research work of Filip Zoric is supported by the Croatian Science Foundation under the project “Young Researchers’ Career Development Project – Training New Doctoral Students” (DOK-2020-01).



\bibliographystyle{ieeetr}
\bibliography{bibliography/RAL_omco}

\end{document}